\documentclass[letterpaper, 10 pt, conference]{ieeeconf}  

\IEEEoverridecommandlockouts                              

\usepackage{amsmath}
\usepackage{flushend}
\usepackage{url}
\usepackage{microtype}
\usepackage{graphicx}
\usepackage{subcaption}   
\usepackage{booktabs} 
\usepackage{wrapfig}
\usepackage{amssymb}
\usepackage{mathtools}
\usepackage{amsfonts}       
\usepackage[capitalize,nameinlink]{cleveref}        
\usepackage{tabularx}
\usepackage{makecell}
\usepackage{wasysym}
\usepackage{multirow}
\usepackage{array}
\usepackage[usenames,dvipsnames,table,xcdraw]{xcolor}
\usepackage[bottom]{footmisc}
\usepackage{threeparttable}
\crefformat{figure}{Fig.~#2#1#3} 
\crefname{figure}{Fig.}{Figs.}
\crefname{subfigure}{Fig.}{Figs.}

\title{\LARGE \bf
WAMJET: A Harness for World Action Model Acceleration
}

\author{Le Chen$^{1}$, Lixin Liu$^{2,\dagger}$, Jan Schneider$^{1}$, Zeju Qiu$^{1}$, Simon Guist$^{1}$, Bernhard Schölkopf$^{1}$, Dieter Büchler$^{1,3}$
\thanks{$^1$Le Chen, Jan Schneider, Zeju Qiu, Simon Guist, Bernhard Schölkopf, and Dieter Büchler are with Max Planck Institute for Intelligent Systems, Tubingen, Germany.}%
\thanks{$^2$Lixin Liu is with The Chinese University of Hong Kong, Hong Kong SAR, China.}%
\thanks{$^3$Dieter Büchler is also with Johannes Kepler University Linz, Austria.}%
\thanks{$^{\dagger}$Corresponding Author: Lixin Liu (lxliu@cse.cuhk.edu.hk).}%
\thanks{The code is available at \protect\url{https://github.com/liulixinkerry/WAMJET}.}
}

\begin{document}
\bstctlcite{IEEEexample:BSTcontrol}
\maketitle
\thispagestyle{empty}
\pagestyle{empty}
\begin{abstract}
World Action Models (WAMs) leverage pretrained video foundation models for robot manipulation, but their large backbones and video-action co-prediction are expensive. Although existing acceleration techniques offer many ways to reduce this cost, selecting and composing them requires substantial engineering for each model and hardware platform. To tackle this bottleneck, we present WAMJET, an agentic harness that accelerates WAM inference by equipping coding agents with reusable optimization guidance and measurement and validation tools. WAMJET follows a bottleneck-driven workflow where the agent profiles inference, modifies targeted code, validates effects, and iteratively refines the acceleration stack as bottlenecks shift, while preserving action quality. Experiments span six WAMs, three coding agents, and two GPU architectures. WAMJET achieves up to 9.95× lossless speedup over upstream implementations. Approximation and hardware-aware optimization yield additional latency reductions, with comparable success rates. The results show that WAMJET can produce effective acceleration stacks for WAM deployment.

\end{abstract}
\section{Introduction}
\label{sec:introduction}

Building generalist policies requires not only mapping visual observations to actions but also understanding and predicting the consequences of actions in diverse environments.
Video generation models learn representations of such dynamics from large-scale video datasets.
Many of these videos contain intentional behavior like object manipulation and scene changes because someone or something acted.
That makes video models attractive backbones for generalist manipulation. 
Before seeing any robot actions, the backbone already encodes useful links between language, visual change, and plausible object interactions. 
World Action Models (WAMs) build on these models and extend them with action-generation capabilities.
They inherit priors from video pretraining and couple action generation with video-based world modeling for more robust robot manipulation across diverse environments.
Compared with Vision-Language-Action Models (VLAs), WAMs are appealing because predicting the future states of the environment may help capture better physical dynamics and temporal information. 

Modern video generation models are often billion-scale parameter models and continue to scale toward higher generation quality and higher fidelity. 
This scaling provides stronger priors about how visual scenes evolve but also increases inference cost. 
The same backbones that make WAMs attractive also make them expensive to run. Inference may involve sampling a large video model, several denoising steps, and repeated vision-language processing. Beyond those computations, data movement, synchronization, and cache updates can also add substantial latency. 
This computational burden poses a significant challenge for real-time control in robotics, where timely decision-making is crucial.

Many techniques can accelerate inference, including compilation, kernel fusion, caching, and quantization. The difficulty is deciding which techniques to apply and how to combine them. A model dominated by repeated conditioning needs different optimizations from one governed by attention or memory transfers. Hardware also affects which kernels or data types are efficient. Moreover, optimizations can interact, which means removing one bottleneck can expose another, and a change that initially gives little benefit may become useful later. Developing an effective acceleration strategy, therefore, requires repeated profiling, implementation, and measurement for each deployment.
Preserving the policy's behavior adds another challenge. Reusing a prompt embedding while the instruction is unchanged can avoid redundant computation. Reusing visual features after the observation changes, however, approximates the policy's input. Quantization can introduce further error, whose effect on manipulation may be difficult to predict from tensor differences alone. An optimization process must account for both inference speed and the effect of its changes on robot actions.

\begin{figure*}[t!]
    \centering
    \includegraphics[width=0.99\textwidth]{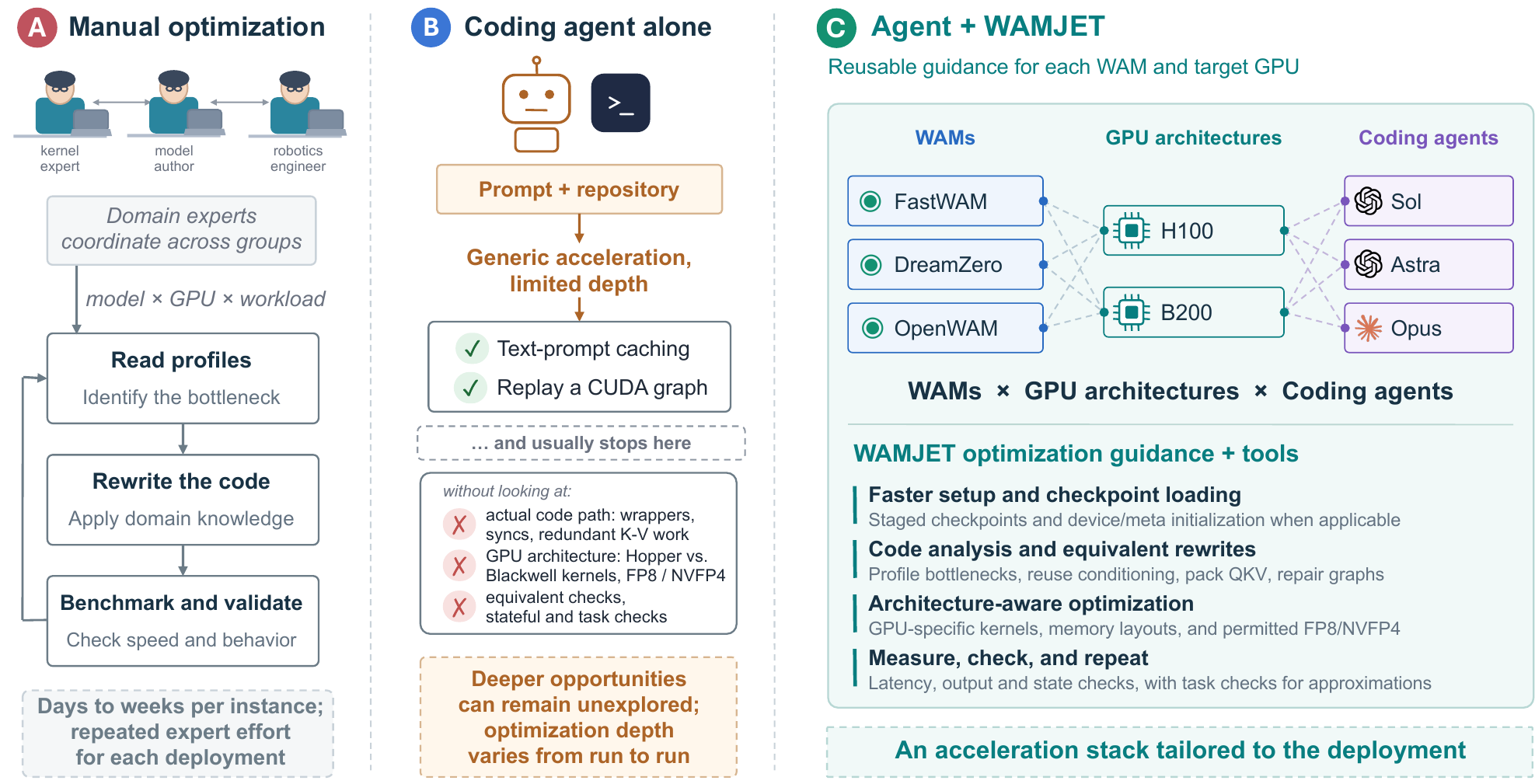}
    \vspace{-0.2em}
    \caption{Overview of WAMJET. (A) Manual optimization requires diverse domain expertise, with substantial engineering for each deployment. (B) A coding agent without guidance may only apply generic optimizations while leaving deeper optimization opportunities unexplored. (C) WAMJET equips the agent with reusable guidance and tools for faster startup, bottleneck analysis, hardware-aware optimization, and iterative validation, supporting acceleration strategies tailored to different WAMs and GPU architectures.}
    \label{fig:fig1_harness_overview}
    \vspace{-0.4cm}
\end{figure*}

Accelerating WAM inference is not a one-time optimization but requires a recipe that composes techniques for a specific model, hardware stack, and task. The best recipe for a certain WAM may not directly transfer to other models. As illustrated in Fig.~\ref{fig:fig1_harness_overview}, hand-crafting an acceleration strategy that works well on a specific model and hardware setup requires substantial engineering effort and may not generalize to other scenarios.
LLM coding agents offer a way to automate this process, but reliable optimization over many iterations requires clear constraints, appropriate tools, and feedback on both latency and correctness. Harness engineering provides this structure through agent-readable workflows, measurement tools, and explicit validation criteria~\cite{lee2026meta}. This shifts human effort toward designing reusable guidance, while the agent conducts experiments, assesses their outcomes, and uses the evidence to guide subsequent optimizations.

In this work, we present WAMJET, an agentic harness for accelerating WAM inference. WAMJET equips coding agents with optimization guidance and tools for profiling, latency measurement, numerical comparison, and paired task analysis. Starting from a runnable policy, the agent identifies bottlenecks, implements lossless or approximate accelerations, and benchmarks and validates each candidate. It retains useful changes and repeats the process within a specified budget. Throughout the search, the agent adapts its optimization choices to the target hardware, records experiment outcomes, and stores reusable findings in external memory. This iterative process produces acceleration stacks tailored to the target policy and hardware.

Our contributions can be summarized in several aspects:
\begin{itemize}
    \item We introduce WAMJET, a harness that combines reusable optimization guidance with measurement and validation tools for accelerating WAM inference.
    \item We develop a bottleneck-driven workflow that iteratively composes acceleration techniques and checks their effects on outputs, recurrent state, and task performance.
    \item We evaluate WAMJET across different WAMs and GPU architectures, examining the optimizations found by different agent campaigns, the benefit of further search, and the gains and limitations of approximate inference. Extensive experiments demonstrate that our training-free framework can achieve significant acceleration.
\end{itemize}

\begin{figure*}[t!]
    \centering
    \includegraphics[width=0.99\textwidth]{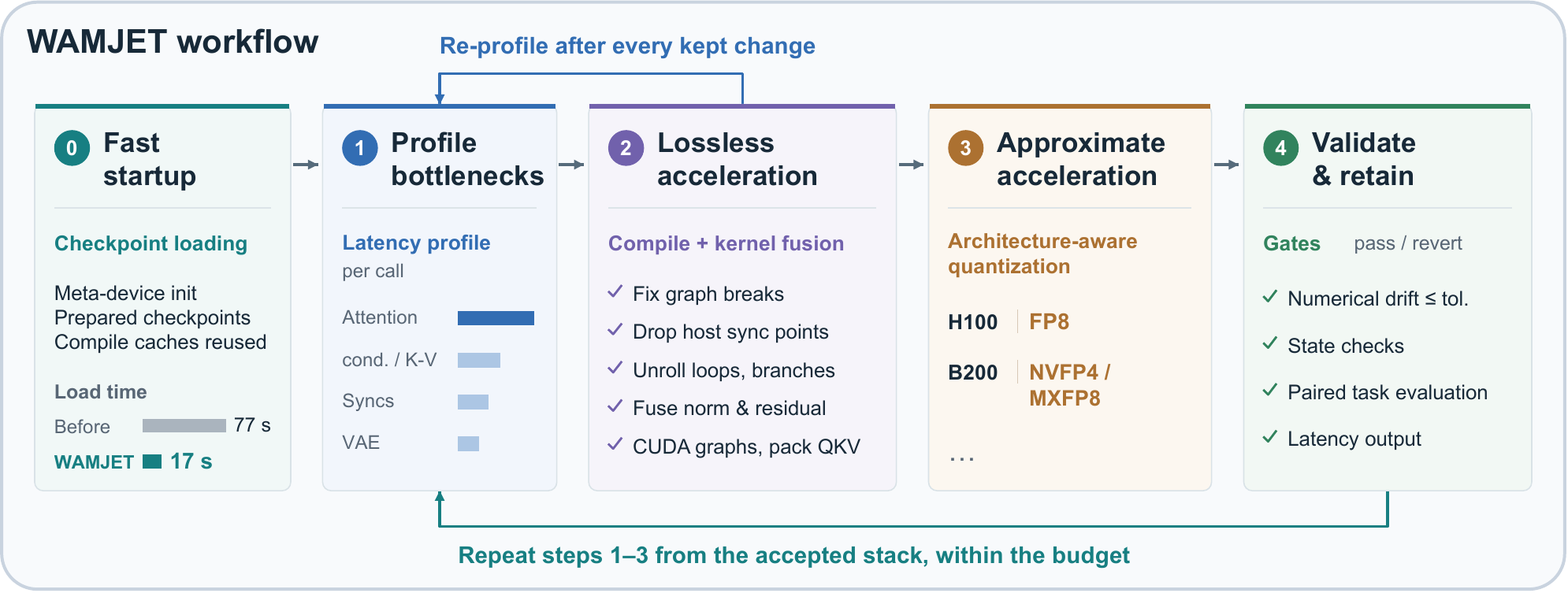}
    \vspace{-0.2em}
    \caption{WAMJET's bottleneck-driven optimization workflow. The agent reduces startup costs, profiles inference to identify bottlenecks, explores lossless acceleration followed by approximate techniques when permitted, and validates candidates. The agent retains accepted changes, reassesses the remaining bottlenecks, and continues searching within the specified budget.} 
    \label{fig:fig2_loop}
    \vspace{-0.4cm}
\end{figure*}
\section{Related Work}
\label{sec:related_work}

\subsection{Foundation Models for Robot Manipulation}

Foundation models support robot manipulation through both direct action prediction and higher-level decision making.
RT-2 transfers vision-language pretraining into control by representing robot actions as tokens, while OpenVLA provides an open-source vision-language-action (VLA) model trained on diverse robot demonstrations~\cite{brohan2023rt2,kim2024openvla}.
$\pi_0$ combines a pretrained vision-language backbone with flow-matching action generation, and Octo learns a generalist diffusion policy from heterogeneous robot data~\cite{black2024pi0,octo2024}.
Other approaches leverage language models: SayCan selects skills based on learned affordances~\cite{ahn2022saycan}, while Code as Policies generates executable programs for robot control~\cite{liang2022codeaspolicies}.

A related line connects policy learning with future visual prediction.
UniPi formulates planning as text-conditioned video generation and extracts executable actions from the resulting visual plans~\cite{du2023unipi}.
Unified models couple video and action learning: UVA learns a shared latent representation with decoupled diffusion heads, while UWM integrates video and action diffusion in a single transformer with independently controlled diffusion timesteps~\cite{li2025uva,zhu2025uwm}.
DreamZero builds on a pretrained video diffusion backbone to jointly predict future observations and robot actions~\cite{ye2026dreamzero}.
WAMJET accelerates the inference code of a trained policy, with numerical and task evaluations used to assess the effect of each optimization.

\subsection{Accelerating Video Generation and World Action Models}

Diffusion acceleration operates at several levels of the inference computation.
DPM-Solver reduces the number of model evaluations through higher-order numerical integration, whereas progressive distillation trains a student to reproduce a longer sampling process with fewer steps~\cite{lu2022dpmsolver,salimans2022distillation}.
Training-free caching methods instead exploit redundancy along the denoising trajectory: TeaCache uses timestep-conditioned input changes to guide reuse, and TaylorSeer forecasts intermediate features rather than simply reusing them~\cite{liu2025teacache,liu2025taylorseer}.
Sparse VideoGen reduces attention computation through spatial-temporal sparsity, token merging reduces redundant token processing, and SVDQuant enables low-bit diffusion inference with a high-precision low-rank branch~\cite{xi2025sparsevideogen,bolya2023tomesd,li2024svdquant}.
These approximations differ from implementations such as FlashAttention, which accelerates exact dense attention through IO-aware tiling and reduced memory traffic~\cite{dao2022flashattention}.

WAM-specific works explore both architectural efficiency and deployment-time optimization.
DreamZero combines caching, quantization, compilation, kernel optimization, and asynchronous execution in a model-specific control stack~\cite{ye2026dreamzero}.
Fast-WAM retains video co-training but removes test-time future-video generation~\cite{yuan2026fastwam}.
Light-WAM further adopts a compact video backbone and state-fusion action decoding, predicting action chunks without a generative action expert~\cite{li2026lightwam}.
These methods require retraining and thus are not drop-in optimizations for arbitrary checkpoints.
In contrast, C$^3$ache is a training-free WAM accelerator that reuses action-expert residuals across consecutive inference chunks, exploiting cross-chunk redundancy~\cite{zhao2026c3ache}.

While these methods offer complementary acceleration mechanisms, they differ in the evaluation criteria: numerical equivalence, perceptual similarity, and task success are not interchangeable, e.g., video fidelity alone does not ensure manipulation performance.
WAMJET builds on these techniques, selecting and combining them according to the bottlenecks of each model and hardware platform.

\subsection{Agentic Workflows and Harness Engineering}

Agentic systems make iterative tool use and external feedback central to problem-solving. ReAct introduced the reasoning–action loop~\cite{yao2022react}, SWE-agent built an agent–computer interface for solving software engineering tasks~\cite{yang2024sweagent}, and agent harnesses go further by designing the whole runtime system around the agent~\cite{zhong2026harness}.
In performance engineering, CUDA-LLM and CudaForge use executable correctness checks and hardware-level performance measurements to iteratively improve GPU kernels~\cite{chen2025cudallm,zhang2025cudaforge}.

Robotics harnesses apply these ideas to different stages of the robot-learning lifecycle.
HARBOR automates robot RL training by coordinating specialized agents from environment setup to reward tuning and to policy training~\cite{li2026harbor}.
Nautilus standardizes policy-benchmark-robot integration through typed contracts and validated workflows~\cite{jin2026nautilus}.
Show-Harness instead operates within the embodied control loop: it exposes compact semantic actions to a VLM and deterministically maps them into robot motion, with execution feedback from observations~\cite{chen2026showharness}.
WAMJET instead applies agentic optimization to the inference code of a trained WAM.

The most related framework is the Sol Video Inference Engine~\cite{li2026sol}, which optimizes video generation by assigning caching, sparse attention, token pruning, quantization, and kernel fusion to parallel agents, then integrating results using human quality feedback.
WAMJET addresses the additional requirements of action generation: optimizations must preserve the policy's recurrent state, respond correctly to changing observations, and be evaluated for their effect on manipulation.
Our workflow combines bottleneck analysis with numerical and stateful checks and uses paired task evaluation to assess approximations.

\section{Preliminaries}
\label{sec:preliminaries}

A visuomotor policy maps current observations plus a goal or instruction to robot actions. A world model predicts future visual or latent states from the current state plus an action or goal abstraction. A WAM sits at the overlap: it leverages a pretrained video generation world model backbone as a prior and predicts both future states and robot actions.

We consider robot manipulation from visual observations and language instructions. At time step $t$, let $x_t=(o_t,q_t,l)$ denote the visual observation, proprioceptive state, and language instruction. A standard policy models the conditional distribution over an action chunk $\mathbf{a}_t=(a_t,\ldots,a_{t+H-1})$ of horizon \(H\) given $x_t$: 
\begin{equation}
    \mathbf{a}_t \sim p_\theta(\mathbf{a}_t\mid x_t)
\end{equation}
where $\theta$ denotes the policy parameters.

WAMs learn to predict robot actions alongside future states. 
Let $\mathbf{s}_t=(s_{t+1},\ldots,s_{t+T})$ denote future video frames or their latent representations over a horizon $T$. WAMs jointly generate both future states and actions:
\begin{equation}
    (\mathbf{s}_t,\mathbf{a}_t)\sim
    p_\theta(\mathbf{s}_t,\mathbf{a}_t\mid x_t).
\end{equation}
The joint distribution can be factorized as
\begin{equation}
\begin{split}
    p_\theta(\mathbf{s}_t,\mathbf{a}_t\mid x_t)
    =p_\theta(\mathbf{s}_t\mid x_t)\cdot p_\theta(\mathbf{a}_t\mid x_t,\mathbf{s}_t)
\end{split}
\end{equation}
where future states are first imagined by the video backbone, and actions are then inferred conditioned on the imagined future, so that the action distribution effectively marginalizes over imagined futures rather than mapping the current context to actions directly. In practice, the inference procedure depends on the architecture; this factorization is instantiated either by jointly denoising future frames and actions within a shared model~\cite{ye2026dreamzero}, or by first generating future video and then feeding it to a separate inverse dynamics module~\cite{li2026lingbot,yuan2026fastwam}. The expensive stages, therefore, differ across WAMs, even when they share the same video model backbone. 
The dominant cost may lie in visual encoding, repeated denoising, action generation, or the transfers and synchronization between these stages. It depends on the architecture and deployment platform, so a technique that helps one WAM might not help another.
%

\section{WAMJET}
\label{sec:wamjet}

\subsection{Overview}

WAMJET is a harness that guides a coding agent in accelerating a trained WAM policy. Its optimization skill describes how to establish a baseline, identify bottlenecks, test changes, and validate results. Python tools support this workflow through inference timing, trace analysis, numerical comparisons, and paired task analysis. Table~\ref{tab:wamjet-harness-components-one-column} summarizes these components by functional role.
The user specifies the policy code and checkpoint, inference configuration, quality requirements, and search budget. Within these constraints, the agent searches for a faster implementation, recording each experiment and its accept/revert decision. It returns the selected implementation with the measurements and validation evidence needed to reproduce and assess the result.

As demonstrated in Fig.~\ref{fig:fig2_loop}, WAMJET guides the agent through an iterative bottleneck-driven optimization workflow. The agent first establishes a runnable baseline and reduces startup costs to allow more experiments within the budget. It then profiles inference to identify bottlenecks, explores corresponding lossless acceleration techniques, and benchmarks each candidate under comparable conditions. Experiment records guide subsequent trials, while reusable findings are retained in external memory. When approximation is permitted, the agent also explores approximate techniques, accepting candidates only after the required checks establish that they satisfy the user's quality requirements. After each change, the agent reassesses the remaining bottlenecks and continues searching until it meets the user requirements or reaches the budget limit. 

\begin{table}[t]
\centering
\caption{Functional components of the WAMJET harness.}
\vspace{-0.5em}
\label{tab:wamjet-harness-components-one-column}
\begingroup
\newcommand{\WamjetComponent}[1]{{\fontsize{7.2}{8.6}\selectfont\ttfamily\bfseries #1}}
\definecolor{WamjetRowRule}{HTML}{8FA5AA}
\definecolor{Wamjet0Band}{HTML}{DDECEE}
\definecolor{Wamjet0Body}{HTML}{F3F6F7}
\definecolor{Wamjet1Band}{HTML}{DDECEE}
\definecolor{Wamjet1Body}{HTML}{F3F6F7}
\definecolor{Wamjet2Band}{HTML}{DDECEE}
\definecolor{Wamjet2Body}{HTML}{F3F6F7}
\fontsize{8}{9.5}\selectfont
\setlength{\tabcolsep}{3pt}
\renewcommand{\arraystretch}{1.12}
\setlength{\extrarowheight}{0.6pt}
\setlength{\aboverulesep}{2.2pt}
\setlength{\belowrulesep}{2.2pt}
\begin{tabularx}{\linewidth}{>{\raggedright\arraybackslash}p{0.30\linewidth}>{\raggedright\arraybackslash}X}
\toprule
\textbf{Component} & \textbf{Function} \\
\midrule
\rowcolor{Wamjet0Band} \multicolumn{2}{l}{\strut\textit{Agent guidance}} \\
\rowcolor{Wamjet0Body} \WamjetComponent{Run configuration} & Defines policy, checkpoint, workload, quality requirements, and search budget limits. \\
\arrayrulecolor{WamjetRowRule}\specialrule{0.5pt}{0pt}{0pt}\arrayrulecolor{black}
\rowcolor{Wamjet0Body} \WamjetComponent{Optimization skill} & Provides procedures for bottleneck analysis, code changes, accept/revert decisions, and validation. \\
\arrayrulecolor{WamjetRowRule}\specialrule{0.5pt}{0pt}{0pt}\arrayrulecolor{black}
\rowcolor{Wamjet0Body} \WamjetComponent{External memory} & Records validated notes for reuse across optimization runs. \\
\noalign{\vskip 3pt}
\rowcolor{Wamjet1Band} \multicolumn{2}{l}{\strut\textit{Execution and Analysis}} \\
\rowcolor{Wamjet1Body} \WamjetComponent{Policy runner} & Runs the policy with reference checks, timing measurement, and run records. \\
\arrayrulecolor{WamjetRowRule}\specialrule{0.5pt}{0pt}{0pt}\arrayrulecolor{black}
\rowcolor{Wamjet1Body} \WamjetComponent{Environment preflight} & Checks package dependencies, software/GPU compatibility, and inference execution. \\
\arrayrulecolor{WamjetRowRule}\specialrule{0.5pt}{0pt}{0pt}\arrayrulecolor{black}
\rowcolor{Wamjet1Body} \WamjetComponent{Endpoint timing} & Measures synchronized warmed inference latency. \\
\arrayrulecolor{WamjetRowRule}\specialrule{0.5pt}{0pt}{0pt}\arrayrulecolor{black}
\rowcolor{Wamjet1Body} \WamjetComponent{Trace analysis} & Locates GPU hotspots, synchronization, memory traffic, compilation, and fusion opportunities. \\
\arrayrulecolor{WamjetRowRule}\specialrule{0.5pt}{0pt}{0pt}\arrayrulecolor{black}
\rowcolor{Wamjet1Body} \WamjetComponent{Startup analysis} & Measures loading phases and checkpoint page-cache residency. \\
\arrayrulecolor{WamjetRowRule}\specialrule{0.5pt}{0pt}{0pt}\arrayrulecolor{black}
\rowcolor{Wamjet1Body} \WamjetComponent{Deadline analysis} & Compares inference latency with the available replanning interval. \\
\noalign{\vskip 3pt}
\rowcolor{Wamjet2Band} \multicolumn{2}{l}{\strut\textit{Validation}} \\
\rowcolor{Wamjet2Body} \WamjetComponent{Output checks} & Checks output structure, finiteness, input identity, and numerical drift. \\
\arrayrulecolor{WamjetRowRule}\specialrule{0.5pt}{0pt}{0pt}\arrayrulecolor{black}
\rowcolor{Wamjet2Body} \WamjetComponent{State checks} & Replays changed inputs against saved references to detect stale outputs. \\
\arrayrulecolor{WamjetRowRule}\specialrule{0.5pt}{0pt}{0pt}\arrayrulecolor{black}
\rowcolor{Wamjet2Body} \WamjetComponent{Paired task analysis} & Compares matched episode outcomes and reports success differences and paired statistics. \\
\bottomrule
\end{tabularx}
\par\vspace{3pt}
\endgroup
\vspace{-0.6cm}
\end{table}

\subsection{Agent-Native Acceleration Workflow}
\label{sec:workflow}

\noindent\textbf{Establish a baseline.}
The agent uses environment preflight checks to establish a runnable inference call on the target GPU and resolves any compatibility issues. The same fixes are applied to both the baseline and subsequent candidates. It then saves the original outputs and measures latency over repeated calls after warmup, keeping the checkpoint, inputs, random seeds, and inference settings fixed. Repeated baseline outputs can also reveal numerical variation that should be considered when checking later changes. Reusing the environment, local checkpoints, and compiler caches reduces the time spent on setup. When loading limits iteration, startup analysis measures loading phases and checkpoint page-cache residency so comparisons use comparable cache states.

\noindent\textbf{Find the bottleneck.}
The agent inspects the inference code and profiles the stages that appear expensive. Trace analysis helps locate costly modules, repeated operations, memory transfers, and synchronization. It then chooses an optimization that addresses the measured bottleneck using the techniques in Section~\ref{sec:mechanisms}. 
As tracing can substantially slow execution, latency is measured without the profiler. 
The inference timing tool synchronizes the GPU, retains the latency series, and checks whether the measurements have reached a steady level.

\noindent\textbf{Test the change.}
The agent measures the modified implementation on the same GPU and compares its outputs with the saved baseline. Cache and CUDA-graph changes are also tested with changing inputs and resets. 
Changes that make inference slower, produce invalid outputs, or violate the required state or quality constraints are reverted. Unexpected numerical differences are investigated using the criteria in Section~\ref{sec:correctness}.
A correct change with unchanged measured latency may be retained if it enables a specific follow-up optimization. Such changes are recorded as having no measured speedup.
Approximate candidates can be explored before task evaluation is available; their acceptance remains pending until that evaluation is complete.

\noindent\textbf{Repeat and validate.}
After an improvement, the agent examines the remaining computation and decides the next experiment. This step matters because the benefit of an optimization depends on the changes already applied. Search continues within the time budget, with time reserved to remeasure the original policy and the best candidate under comparable conditions. The final evaluation checks numerical outputs, stateful behavior, and task performance as required by the type of change. Loading and compilation costs are recorded separately from warmed inference latency.
Deadline analysis relates inference latency to the available replanning interval. For $K$ actions executed between replans at control frequency $f$, this interval is $K/f$. This reports deployment headroom while keeping the workload fixed.

External memory records validated notes from earlier runs, such as environment fixes and model-specific findings. The agent reuses these notes and can incorporate broader lessons into the optimization skill after a run. In Sec.~\ref{sec:experiments}, when evaluating the harness itself, the guidance is held fixed, and both settings receive the same resources, search budget, and access to prior information.

\subsection{Correctness and Candidate Acceptance}
\label{sec:correctness}
Different optimizations require different checks. WAMJET distinguishes three tiers:
\begin{itemize}
    \item \textbf{Tier A: lossless acceleration.}
    The change preserves the mathematical computation, precision, workload, and state behavior. The agent assesses eligibility through mathematical and code review and focused output and state checks. Equivalent compilation, fusion, and changes to GEMM or reduction order may introduce floating-point differences. Unexpected drift is investigated rather than rejected solely for exceeding an earlier numerical tolerance.
    
    \item \textbf{Tier B: validated approximation.} The change approximates internal computation while keeping the denoising steps, schedule, action horizon, and required outputs fixed. It must pass paired task evaluation within a specified bound on quality loss.
    \item \textbf{Tier C: changed workload or unvalidated approximation.} This includes fewer denoising steps, changed horizons or schedules, and approximations that lack passing task evaluation.
\end{itemize}
The user sets the maximum acceptable tier. Invalid outputs, stale state, and known task-quality failures are rejected in all tiers. Lossless acceleration does not require bit-identical outputs.
The numerical checker verifies output keys, shapes, dtypes, finite values, and input identity against the saved reference. For each tensor, it reports the maximum absolute error and relative $\ell_2$ error. These measurements are assessed against the saved tolerances and, when available, the baseline's run-to-run variation. The original reference is retained throughout the search so that successive small changes cannot hide cumulative drift. However, we find that the numerical drift checker could possibly make the acceleration too conservative, so in practice, the numerical checker is usually turned off.

The agent tests stateful optimizations using input sequences such as $A\!\rightarrow\!B\!\rightarrow\!A$, comparing each output with its baseline reference under the same state history. It also verifies that resets restore the intended state. These checks help detect stale caches and incorrect reuse of buffer contents from earlier calls.
Task performance is evaluated by running the baseline and candidate on matched episodes with identical initial states and random seeds. The number of episodes and acceptable performance loss are specified in advance, and final evaluation uses episodes excluded from calibration and optimization. WAMJET reports the difference in success rate and applies an exact McNemar test to the paired outcomes. Acceptance requires evidence that the candidate meets the predefined quality criterion.

\subsection{Acceleration Techniques}
\label{sec:mechanisms}
The bottleneck determines which techniques the agent tries first. When inference repeatedly computes the same conditioning, prompt embeddings or K/V projections can be reused as long as their inputs remain unchanged. Unnecessary synchronization and distributed wrappers can be removed, while static buffers and CUDA graphs reduce allocation and dispatch overhead. These changes must preserve cache updates and ensure that graph replay receives the current inputs.

When most time is spent in GPU kernels or memory transfers, the agent can try packed projections, faster implementations of the same attention operation, different tensor layouts, or compiler specialization. Fusing normalization, modulation, activation, and residual operations can reduce intermediate memory traffic. Each change is evaluated through its effect on the complete inference call, since a faster kernel may have little effect on overall latency.

When approximation is allowed, the search can also explore quantization and approximate reuse of features or K/V tensors. The agent verifies that the intended low-precision or reuse path executes and measures the resulting error. Conversion and cache-maintenance costs can offset the expected speedup, so the benefit must be measured for the actual workload. Useful candidates are then evaluated on robotic tasks. As earlier optimizations change the bottleneck, the agent can revisit techniques that initially gave little benefit.

\section{Experiments}
\label{sec:experiments}
\begin{table*}[t]
\centering
\caption{Latency on H100 across baseline, w/o WAMJET, and w/ WAMJET Lossless across LLMs and WAMs.}
\vspace{-0.5em}
\setlength{\tabcolsep}{3pt}
\resizebox{0.98\textwidth}{!} 
{
\begin{tabular}{c|c|c|cc|cccc}
\toprule
\multirow{2}{*}{LLM}           & \multirow{2}{*}{WAM} & Baseline     & \multicolumn{2}{c|}{w/o WAMJET} & \multicolumn{4}{c}{w/ WAMJET Lossless (Ours)}                                                           \\
                               &                      & Latency (ms) & Latency (ms)      & Speedup     & $R_1$ Latency (ms)$\ast$ & $R_2$ Latency (ms)$\ast$ & $R_1$/$R_2$$\dagger$ & Speedup$\ddagger$ \\ \midrule
\multirow{4}{*}{GPT-5.6-sol}   & Fast-WAM              & 84.7         & 59.2              & 1.43x      & 50.4                     & 37.4                     & 1.35x                & \textbf{2.27x}    \\
                               & Cosmos-Policy        & 384.2        & 369.2             & 1.04x      & 236.2                    & 163.1                    & 1.45x                & \textbf{2.36x}    \\
                               & DreamZero            & 2780.7       & 2687.7            & 1.03x      & 2373.9                   & 2030.6                   & 1.17x                & \textbf{1.37x}    \\
                               & LingBot-VA           & 5323.6       & 2670.2            & 1.99x      & 755.8                    & 686.5                    & 1.10x                & \textbf{7.75x}    \\ \midrule
\multirow{4}{*}{Claude-opus-5} & Fast-WAM              & 84.7         & 54.1              & 1.57x      & 47.5                     & 43.1                     & 1.10x                & \textbf{1.96x}    \\
                               & Cosmos-Policy        & 384.2        & 263.7             & 1.46x      & 143.0                    & 134.9                    & 1.06x                & \textbf{2.85x}    \\
                               & DreamZero            & 2780.7       & 2454.1            & 1.13x      & 1948.7                   & 1904.5                   & 1.02x                & \textbf{1.46x}    \\
                               & LingBot-VA           & 5323.6       & 894.4             & 5.95x      & 902.2                    & 537.5                    & 1.68x                & \textbf{9.91x}    \\ \midrule
\multirow{4}{*}{GPT-6-astra}   & Fast-WAM              & 84.7         & 59.2              & 1.43x      & 47.5                     & 34.9                     & 1.36x                & \textbf{2.43x}    \\
                               & Cosmos-Policy        & 384.2        & 264.6             & 1.45x      & 164.8                    & 131.9                    & 1.25x                & \textbf{2.91x}    \\
                               & DreamZero            & 2780.7       & 2502.9            & 1.11x      & 1938.2                   & 1785.2                   & 1.09x                & \textbf{1.56x}    \\
                               & LingBot-VA           & 5323.6       & 966.7             & 5.51x      & 634.9                    & 534.9                    & 1.19x                & \textbf{9.95x}    \\ \bottomrule
\end{tabular}
}
\begin{tablenotes}
\item  All accelerations reported in this table are lossless, with no statistically significant difference in policy performance.
\item  $\ast$ $R_x$ denotes executing the WAMJET optimization for $x$ rounds. Our final reported results use the $R_2$ latency.
\item  $\dagger$ $R_1$/$R_2$ denotes the speedup of the second round $R_2$ result over the first round $R_1$ result.
\item  $\ddagger$ Total speedup denotes the speedup of our final $R_2$ result over the baseline.
\end{tablenotes}
\label{table:exp:wamjet_lossless_round}
\vspace{-0.25cm}
\end{table*}
\begin{table*}[t]
\centering
\caption{Latency of two WAMs under baseline and different WAMJET acceleration levels on two GPU architectures.}
\vspace{-0.5em}
\setlength{\tabcolsep}{7pt}
\resizebox{0.98\textwidth}{!} 
{
\begin{tabular}{c|c|c|cc|ccc}
\toprule
\multirow{2}{*}{WAM}      & \multirow{2}{*}{GPU} & Baseline     & \multicolumn{2}{c|}{w/ WAMJET Lossless} & \multicolumn{3}{c}{w/ WAMJET Approx.}        \\
                          &                      & Latency (ms) & Latency (ms)          & Speedup         & Latency (ms) & Speedup & Approx. Method      \\ \midrule
DreamZero                 & H100                 & 2780.7       & 1785.2                & 1.56x           & 1297.0       & \textbf{2.14x}   &  FP8         \\
DreamZero                 & B200                 & 1598.5       & 823.9                 & 1.94x           & 722.6        & \textbf{2.21x}   & NVFP4 + FP8 \\
Cosmos3-Nano-Policy-DROID & H100                 & 814.8        & 768.8                 & 1.06x           & 651.2        & \textbf{1.25x}   & FP8         \\
Cosmos3-Nano-Policy-DROID & B200                 & 446.3        & 368.3                 & 1.21x           & 304.7        & \textbf{1.47x}   & MXFP8       \\ \bottomrule
\end{tabular}
}
\label{table:exp:wamjet_approx_latency}
\vspace{-0.5cm}
\end{table*}

\subsection{Experimental Setup}
 
We evaluate WAMJET across six WAMs, three coding agents, and two GPU architectures. To assess the harness’s generality, we study lossless acceleration across different combinations of coding agents and WAMs on NVIDIA H100. We further evaluate approximate acceleration on both H100 and B200. As a robustness check, we apply WAMJET to a recently released state-of-the-art WAM on B200 to examine whether it can further accelerate an already optimized inference pipeline.
Inference latency is measured after warm-up, excluding model loading, compilation, and CUDA-graph capture. Each experiment follows its recorded timing and state-update protocol.

\subsection{Lossless Acceleration}
\label{sec:lossless_results}

In this section, we evaluate WAMJET’s effectiveness in lossless acceleration and its generality across different combinations of coding agents and WAMs. Since the acceleration here is lossless, the policy performance remains unchanged.
On H100, we evaluate three coding agents: GPT-5.6-sol xhigh (Codex), GPT-6-astra xhigh (Codex), and Claude-opus-5 xhigh (Claude Code). We pair them with four WAMs: Fast-WAM~\cite{yuan2026fastwam}, DreamZero~\cite{ye2026dreamzero}, Cosmos-Policy-Predict2-2B~\cite{kim2026cosmos}, and LingBot-VA~\cite{li2026lingbot}. To assess the benefit of WAMJET beyond the coding agents’ own optimization capabilities, we compare three settings:
\begin{itemize}
    \item \textbf{Baseline:} the official upstream implementation, with any compatibility fixes shared across all settings.
    \item \textbf{Without WAMJET:} the coding agent starts from the baseline and optimizes inference independently.
    \item \textbf{With WAMJET:} the coding agent starts from the baseline and optimizes inference using WAMJET's guidance and tools.
\end{itemize}

\Cref{table:exp:wamjet_lossless_round} reports the first-round ($R_1$) and second-round ($R_2$) WAMJET configurations. From the results, we can see that: (1) \textbf{WAMJET can effectively accelerate inference speed across different coding agents and WAMs combinations.} All WAMJET configurations achieve speedups over the shared model baselines and have lower reported latency than the corresponding runs without WAMJET in all 12 entries. For LingBot-VA, our harness achieves a significant $9.95\times$ speedup with GPT-6-astra, showing that our harness could benefit more from more advanced coding agents. (2) \textbf{More budget leads to better optimization.} Even though the first round has already accelerated the latency quite a lot, the ratio $R_1$/$R_2$ demonstrates that executing WAMJET for a second round can still further speed up the inference.

\subsection{Approximate Acceleration across GPU Architectures}
We evaluate WAMJET with GPT-6-astra on Cosmos3-Nano-Policy-DROID~\cite{agarwal2026cosmos3} and DreamZero~\cite{ye2026dreamzero}, across H100 and B200 GPUs. We further enable approximate acceleration on top off lossless acceleration to quantify acceleration across different WAMs and GPU architectures. Since we apply approximate acceleration here, we also evaluate task success on RoboLab~\cite{yang2026robolab} using 20 tasks with 10 episodes per task to assess whether these speedups preserve task performance. 

\begin{figure*}[t]
    \centering
    \includegraphics[width=0.95\textwidth]{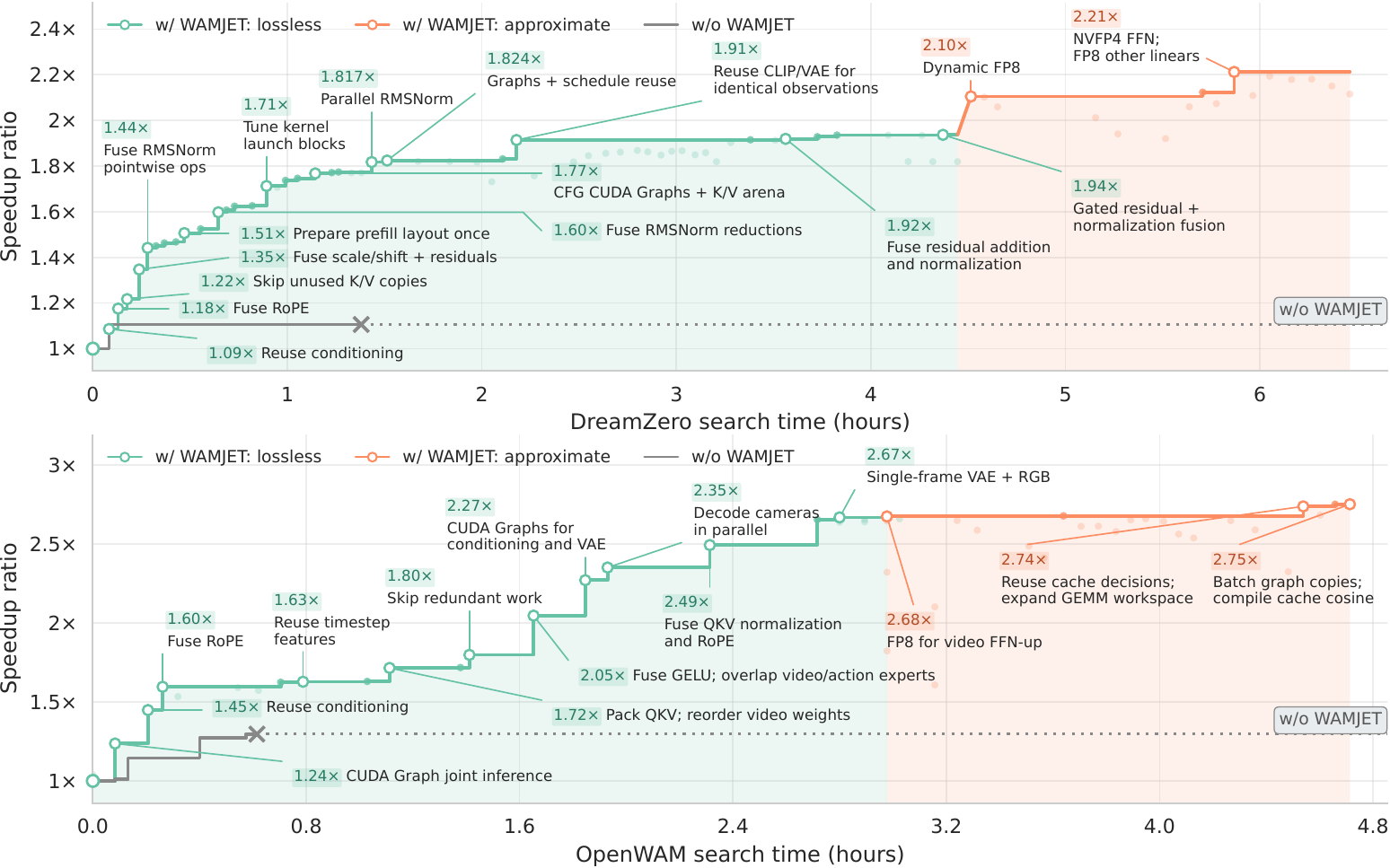}
    \vspace{-0.3em}
    \caption{Optimization progress over search time for DreamZero (top) and OpenWAM (bottom) on B200.}
    \label{fig:wamjet_search}
    \vspace{-0.4cm}
\end{figure*}

As shown in \Cref{table:exp:wamjet_approx_latency}, WAMJET accelerates both policies across H100 and B200. Lossless optimization achieves $1.56\times$ and $1.94\times$ speedups for DreamZero, and $1.06\times$ and $1.21\times$ for Cosmos3, on H100 and B200, respectively. Adding approximate optimization further increases the overall speedups to $2.14\times$ and $2.21\times$ for DreamZero, and $1.25\times$ and $1.47\times$ for Cosmos3, respectively.
These results illustrate the complementary benefits of lossless and approximate optimization. In addition, from Table~\ref{table:exp:wamjet_approx_success} we can see that WAMJET can preserve action quality and task performance even for approximate acceleration, indicating the effectiveness of our iterative validation-involved workflow.

\textbf{WAMJET supports hardware-aware optimization.} The WAMJET-guided agent selects and refines quantization strategies through profiling and empirical search, adapting them to the target model and GPU architecture. On H100, both DreamZero and Cosmos3 use FP8 with per-output-channel weight scaling and per-token activation scaling. On B200, DreamZero uses a mixed-precision configuration: NVFP4 for the feed-forward projections and FP8 for the remaining linear projections within the diffusion-transformer blocks. Cosmos3 instead uses MXFP8 for its generation projections. The agent combines these quantization strategies with lossless optimizations, including compilation and scheduling, and reassesses the remaining bottlenecks after each change.

\subsection{Further Optimization of a Strong WAM Baseline}
\begin{table}[t]
\centering
    \caption{RoboLab subset success rates of Cosmos3-Nano-Policy-DROID and DreamZero under baseline and different WAMJET acceleration levels.}
\vspace{-0.5em}
\setlength{\tabcolsep}{4pt}
\resizebox{0.48\textwidth}{!} 
{
\begin{tabular}{c|c|cc}
\toprule
WAM                                  & Configuration      & Approx. Method      & Success Rate \\ \midrule
\multirow{4}{*}{DreamZero}           & Baseline           & -                   & 25.0\%         \\
                                     & w/ WAMJET Lossless & -                   & 24.0\%         \\
                                     & w/ WAMJET Approx.  &  FP8         & 26.0\%         \\
                                     & w/ WAMJET Approx.  &  NVFP4 + FP8 & 25.0\%         \\ \midrule
\multirow{4}{*}{Cosmos3-Nano-Policy} & Baseline           & -                   & 43.0\%         \\
                                     & w/ WAMJET Lossless & -                   & 47.0\%         \\
                                     & w/ WAMJET Approx.  &  FP8         & 45.5\%         \\
                                     & w/ WAMJET Approx.  &  MXFP8       & 44.5\%         \\ \bottomrule
\end{tabular}
}
\label{table:exp:wamjet_approx_success}
\vspace{-0.5cm}
\end{table}
\begin{table}[t]
\centering
\caption{Latency and success rates for OpenWAM.}
\vspace{-0.5em}
\setlength{\tabcolsep}{2pt}
\resizebox{0.48\textwidth}{!} 
{
\begin{tabular}{c|ccccc}
\toprule
Config             & Precision & Latency (ms) & Speedup & LIBERO  & LIBERO-plus \\ \midrule
Baseline           & BF16      & 87.25        & 1.00x   & 98.95\% & 69.27\%     \\
w/ WAMJET Lossless & BF16      & 32.69        & 2.67x   & 99.10\% & 69.24\%     \\
w/ WAMJET Approx.  & FP8       & 31.70        & 2.75x   & 99.25\% & 69.30\%     \\ \bottomrule
\end{tabular}
}
\label{table:exp:wamjet_openwam}
\vspace{-0.5cm}
\end{table}

Wang et al. recently released OpenWAM~\cite{wang2026openwam}, which achieves state-of-the-art performance across several robotic benchmarks. OpenWAM already incorporates several inference optimizations, making it a useful test of whether WAMJET can guide an agent to find further optimization opportunities. We apply WAMJET-guided GPT-6-astra to optimize OpenWAM on B200, and evaluate on full LIBERO~\cite{liu2023libero} and full LIBERO-plus~\cite{fei25libero-plus}.

The results are shown in \Cref{table:exp:wamjet_openwam}.
OpenWAM already enables the PyTorch compiler and velocity cache, achieving a latency of $87.25$\,ms on B200. 
WAMJET further reduces latency by $32.69$\,ms with lossless optimizations and $31.70$\,ms with approximate optimizations, achieving $2.67\times$ and $2.75\times$ speedups, respectively, while preserving action quality.  
The similar speedups of the lossless and approximate methods indicate that quantization provides limited additional benefit in this setting, with most of the gains coming from lossless methods. These results show that the guided agent can identify substantial acceleration opportunities even in an already highly optimized inference pipeline.

\subsection{Optimization Progress over Search Time}

We plot the optimization progress of GPT-6-astra on DreamZero and OpenWAM using B200. \Cref{fig:wamjet_search} shows individual trials and accepted optimization trajectories, illustrating how performance improves through iterative exploration.
With WAMJET, the coding agent autonomously profiles each candidate, identifies new bottlenecks, and iteratively explores optimization opportunities. In contrast, without our harness, coding agents tend to be overly conservative and terminate early, leaving substantial optimization opportunities unexplored.
Furthermore, \Cref{fig:wamjet_search} also annotates representative points with the key optimizations introduced during the search. Although DreamZero and OpenWAM share techniques such as CUDA Graphs and operator fusion, their final optimization recipes differ even on the same GPU. The results indicate that the guided agent adopts adaptive, model-specific optimization strategies according to each model's bottlenecks.

\subsection{Startup Cost Acceleration}
WAMJET also guides the coding agent in optimizing model startup cost. In recorded B200 runs, DreamZero's startup time decreases from $208.1$ to $30.3$\,s, and OpenWAM's decreases from $79.6$ to $8.2$\,s.
\section{Conclusion}
We presented WAMJET, a harness for accelerating WAM inference through iterative code optimization and evaluation. WAMJET guides a coding agent to find bottlenecks, test acceleration techniques, and check their effects on the policy. Experiments across six models show improvements from conditioning reuse, reduced execution overhead, and kernel optimization, with additional gains from continued search. Quantization can further reduce latency, but its effect on task performance requires evaluation. Future work will focus on broader paired task tests and controlled comparisons of the harness under matched search budgets.

\section*{ACKNOWLEDGMENT}
We used AI for editing, grammar checking, coding assistance under human guidance, and to gain inspiration for style and art design for figure color and layout, but not for the final figures or any other technical content in this manuscript.



\bibliographystyle{./IEEEtran} 
\bibliography{./reference.bib}

@IEEEtranBSTCTL{IEEEexample:BSTcontrol,
  CTLuse_forced_etal = "yes",
  CTLmax_names_forced_etal = "6",
  CTLnames_show_etal = "1"
}

@article{li2026lingbot,
  title = {Causal World Modeling for Robot Control},
  author = {Li, Lin and Zhang, Qihang and Luo, Yiming and Yang, Shuai and Wang, Ruilin and Han, Fei and Yu, Mingrui and Gao, Zelin and Xue, Nan and Zhu, Xing and Shen, Yujun and Xu, Yinghao},
  journal = {arXiv preprint arXiv:2601.21998},
  year = {2026}
}

@article{kim2026cosmos,
  title = {{Cosmos Policy}: Fine-Tuning Video Models for Visuomotor Control and Planning},
  author = {Kim, Moo Jin and Gao, Yihuai and Lin, Tsung-Yi and Lin, Yen-Chen and Ge, Yunhao and Lam, Grace and Liang, Percy and Song, Shuran and Liu, Ming-Yu and Finn, Chelsea and Gu, Jinwei},
  journal = {arXiv preprint arXiv:2601.16163},
  year = {2026}
}

@inproceedings{liu2023libero,
  title = {{LIBERO}: Benchmarking Knowledge Transfer for Lifelong Robot Learning},
  author = {Liu, Bo and Zhu, Yifeng and Gao, Chongkai and Feng, Yihao and Liu, Qiang and Zhu, Yuke and Stone, Peter},
  booktitle = {Advances in Neural Information Processing Systems (NeurIPS), Datasets and Benchmarks Track},
  year = {2023}
}

@article{fei25libero-plus,
    title={{LIBERO-Plus: In-depth Robustness} Analysis of Vision-Language-Action Models},
    author={Senyu Fei and Siyin Wang and Junhao Shi and Zihao Dai and Jikun Cai and Pengfang Qian and Li Ji and Xinzhe He and Shiduo Zhang and Zhaoye Fei and Jinlan Fu and Jingjing Gong and Xipeng Qiu},
    journal = {arXiv preprint arXiv:2510.13626},
    year={2025},
}

@inproceedings{du2023unipi,
  title = {Learning Universal Policies via Text-Guided Video Generation},
  author = {Du, Yilun and Yang, Mengjiao and Dai, Bo and Dai, Hanjun and Nachum, Ofir and Tenenbaum, Joshua B. and Schuurmans, Dale and Abbeel, Pieter},
  booktitle = {Advances in Neural Information Processing Systems (NeurIPS)},
  year = {2023}
}

@article{brohan2023rt2,
  title = {{RT-2}: Vision-Language-Action Models Transfer Web Knowledge to Robotic Control},
  author = {Anthony Brohan and Noah Brown and Justice Carbajal and Yevgen Chebotar and Xi Chen and Krzysztof Choromanski and Tianli Ding and Danny Driess and Avinava Dubey and Chelsea Finn and Pete Florence and Chuyuan Fu and Montse Gonzalez Arenas and Keerthana Gopalakrishnan and Kehang Han and Karol Hausman and Alexander Herzog and Jasmine Hsu and Brian Ichter and Alex Irpan and Nikhil Joshi and Ryan Julian and Dmitry Kalashnikov and Yuheng Kuang and Isabel Leal and Lisa Lee and Tsang-Wei Edward Lee and Sergey Levine and Yao Lu and Henryk Michalewski and Igor Mordatch and Karl Pertsch and Kanishka Rao and Krista Reymann and Michael Ryoo and Grecia Salazar and Pannag Sanketi and Pierre Sermanet and Jaspiar Singh and Anikait Singh and Radu Soricut and Huong Tran and Vincent Vanhoucke and Quan Vuong and Ayzaan Wahid and Stefan Welker and Paul Wohlhart and Jialin Wu and Fei Xia and Ted Xiao and Peng Xu and Sichun Xu and Tianhe Yu and Brianna Zitkovich},
  journal = {arXiv preprint arXiv:2307.15818},
  year = {2023},
}

@article{kim2024openvla,
  title = {{OpenVLA}: An Open-Source Vision-Language-Action Model},
  author = {Moo Jin Kim and Karl Pertsch and Siddharth Karamcheti and Ted Xiao and Ashwin Balakrishna and Suraj Nair and Rafael Rafailov and Ethan Foster and Grace Lam and Pannag Sanketi and Quan Vuong and Thomas Kollar and Benjamin Burchfiel and Russ Tedrake and Dorsa Sadigh and Sergey Levine and Percy Liang and Chelsea Finn},
  journal={arXiv preprint arXiv:2406.09246},
  year={2024}
}

@article{black2024pi0,
  title = {$\pi_0$: A Vision-Language-Action Flow Model for General Robot Control},
  author = {Kevin Black and Noah Brown and Danny Driess and Adnan Esmail and Michael Equi and Chelsea Finn and Niccolo Fusai and Lachy Groom and Karol Hausman and Brian Ichter and Szymon Jakubczak and Tim Jones and Liyiming Ke and Sergey Levine and Adrian Li-Bell and Mohith Mothukuri and Suraj Nair and Karl Pertsch and Lucy Xiaoyang Shi and James Tanner and Quan Vuong and Anna Walling and Haohuan Wang and Ury Zhilinsky},
  journal={arXiv preprint arXiv:2410.24164},
  year={2024}
}

@inproceedings{octo2024,
  title = {Octo: An Open-Source Generalist Robot Policy},
  author = {{Octo Model Team} and Dibya Ghosh and Homer Walke and Karl Pertsch and Kevin Black and Oier Mees and Sudeep Dasari and Joey Hejna and Charles Xu and Jianlan Luo and Tobias Kreiman and {You Liang} Tan and Lawrence Yunliang Chen and Pannag Sanketi and Quan Vuong and Ted Xiao and Dorsa Sadigh and Chelsea Finn and Sergey Levine},
    booktitle = {Proceedings of Robotics: Science and Systems},
    year = {2024},
}

@article{ahn2022saycan,
  title = {{Do As I Can, Not As I Say}: Grounding Language in Robotic Affordances},
  author = {Michael Ahn and Anthony Brohan and Noah Brown and Yevgen Chebotar and Omar Cortes and Byron David and Chelsea Finn and Chuyuan Fu and Keerthana Gopalakrishnan and Karol Hausman and Alex Herzog and Daniel Ho and Jasmine Hsu and Julian Ibarz and Brian Ichter and Alex Irpan and Eric Jang and Rosario Jauregui Ruano and Kyle Jeffrey and Sally Jesmonth and Nikhil J. Joshi and Ryan Julian and Dmitry Kalashnikov and Yuheng Kuang and Kuang-Huei Lee and Sergey Levine and Yao Lu and Linda Luu and Carolina Parada and Peter Pastor and Jornell Quiambao and Kanishka Rao and Jarek Rettinghouse and Diego Reyes and Pierre Sermanet and Nicolas Sievers and Clayton Tan and Alexander Toshev and Vincent Vanhoucke and Fei Xia and Ted Xiao and Peng Xu and Sichun Xu and Mengyuan Yan and Andy Zeng},
  journal={arXiv preprint arXiv:2204.01691},
  year={2022}
}

@inproceedings{liang2022codeaspolicies,
  title = {{Code as Policies}: Language Model Programs for Embodied Control},
  author={Liang, Jacky and Huang, Wenlong and Xia, Fei and Xu, Peng and Hausman, Karol and Ichter, Brian and Florence, Pete and Zeng, Andy},
  booktitle={2023 IEEE International conference on robotics and automation (ICRA)},
  pages={9493--9500},
  year={2023},
  organization={IEEE}
}

@article{li2025uva,
  title = {Unified Video Action Model},
  author = {Shuang Li and Yihuai Gao and Dorsa Sadigh and Shuran Song},
  journal = {arXiv preprint arXiv:2503.00200},
  year = {2025},
}

@inproceedings{zhu2025uwm,
    author = {Zhu, Chuning and Yu, Raymond and Feng, Siyuan and Burchfiel, Benjamin and Shah, Paarth and Gupta, Abhishek},
    title = {{Unified World Models}: Coupling Video and Action Diffusion for Pretraining on Large Robotic Datasets},
    booktitle = {Proceedings of Robotics: Science and Systems (RSS)},
    year = {2025},
}

@article{ye2026dreamzero,
  title = {World Action Models are Zero-shot Policies},
  author={Ye, Seonghyeon and Ge, Yunhao and Zheng, Kaiyuan and Gao, Shenyuan and Yu, Sihyun and Kurian, George and Indupuru, Suneel and Tan, You Liang and Zhu, Chuning and Xiang, Jiannan and others},
  journal={arXiv preprint arXiv:2602.15922},
  year={2026}
}

@article{lu2022dpmsolver,
  title = {{DPM-Solver}: A Fast {ODE} Solver for Diffusion Probabilistic Model Sampling in Around 10 Steps},
  author={Lu, Cheng and Zhou, Yuhao and Bao, Fan and Chen, Jianfei and Li, Chongxuan and Zhu, Jun},
  journal={Advances in neural information processing systems},
  year={2022}
}

@article{salimans2022distillation,
  title = {Progressive Distillation for Fast Sampling of Diffusion Models},
  author={Salimans, Tim and Ho, Jonathan},
  journal={arXiv preprint arXiv:2202.00512},
  year={2022}
}

@inproceedings{liu2025teacache,
  title = {Timestep Embedding Tells: It's Time to Cache for Video Diffusion Model},
  author={Liu, Feng and Zhang, Shiwei and Wang, Xiaofeng and Wei, Yujie and Qiu, Haonan and Zhao, Yuzhong and Zhang, Yingya and Ye, Qixiang and Wan, Fang},
  booktitle={2025 IEEE/CVF Conference on Computer Vision and Pattern Recognition (CVPR)},
  pages={7353--7363},
  year={2025},
  organization={IEEE}
}

@inproceedings{liu2025taylorseer,
  title = {From Reusing to Forecasting: Accelerating Diffusion Models with {TaylorSeers}},
  author={Liu, Jiacheng and Zou, Chang and Lyu, Yuanhuiyi and Chen, Junjie and Zhang, Linfeng},
  booktitle={2025 IEEE/CVF International Conference on Computer Vision},
  year={2025},
}

@article{xi2025sparsevideogen,
  title={{Sparse VideoGen}: Accelerating Video Diffusion Transformers with Spatial-Temporal Sparsity},
  author={Xi, Haocheng and Yang, Shuo and Zhao, Yilong and Xu, Chenfeng and Li, Muyang and Li, Xiuyu and Lin, Yujun and Cai, Han and Zhang, Jintao and Li, Dacheng and others},
  journal={arXiv preprint arXiv:2502.01776},
  year={2025}
}

@article{bolya2023tomesd,
  title={Token Merging for Fast Stable Diffusion},
  author={Bolya, Daniel and Hoffman, Judy},
  journal={CVPR Workshop on Efficient Deep Learning for Computer Vision},
  year={2023}
}

@inproceedings{li2024svdquant,
 title={{SVDQuant}: Absorbing Outliers by Low-Rank Components for 4-Bit Diffusion Models},
 author={Li, Muyang and Lin, Yujun and Zhang, Zhekai and Cai, Tianle and Li, Xiuyu and Guo, Junxian and Xie, Enze and Meng, Chenlin and Zhu, Jun-Yan and Han, Song},
booktitle={The Thirteenth International Conference on Learning Representations},
 year={2025}
}

@article{dao2022flashattention,
  title = {{FlashAttention}: Fast and Memory-Efficient Exact Attention with IO-Awareness},
  author={Dao, Tri and Fu, Dan and Ermon, Stefano and Rudra, Atri and R{\'e}, Christopher},
  journal={Advances in neural information processing systems},
  year={2022}
}

@article{yuan2026fastwam,
  title = {{Fast-WAM}: Do World Action Models Need Test-time Future Imagination?},
  author={Yuan, Tianyuan and Dong, Zibin and Liu, Yicheng and Zhao, Hang},
  journal={arXiv preprint arXiv:2603.16666},
  year={2026}
}

@article{li2026lightwam,
  title = {{Light-WAM}: Efficient World Action Models with State-Fusion Action Decoding},
  author={Li, Ziang and Cheng, Dongzhou and Wang, Yibin and Wang, Shiyue and Xu, Xiaoyang and Weng, Lingxuan and Wang, Juan and Wang, Jiaqi},
  journal={arXiv preprint arXiv:2606.08242},
  year={2026}
}

@article{zhao2026c3ache,
  title = {{C$^3$ache}: Accelerating World Action Models with Cross Inference Chunk Cache},
  author={Zhao, Weisen and Nguyen, Lam and Lu, Zhicong and Shang, Yuzhang},
  journal={arXiv preprint arXiv:2606.08962},
  year={2026}
}

@inproceedings{yao2022react,
  title = {{ReAct}: Synergizing Reasoning and Acting in Language Models},
  author = {Yao, Shunyu and Zhao, Jeffrey and Yu, Dian and Du, Nan and Shafran, Izhak and Narasimhan, Karthik and Cao, Yuan},
  booktitle = {International Conference on Learning Representations (ICLR) },
  year = {2023},
  html = {https://arxiv.org/abs/2210.03629},
}

@article{yang2024sweagent,
  title = {{SWE-agent}: Agent-Computer Interfaces Enable Automated Software Engineering},
  author={Yang, John and Jimenez, Carlos and Wettig, Alexander and Lieret, Kilian and Yao, Shunyu and Narasimhan, Karthik and Press, Ofir},
  journal={Advances in Neural Information Processing Systems},
  volume={37},
  pages={50528--50652},
  year={2024}
}

@article{zhong2026harness,
  title = {{AI Harness Engineering}: A Runtime Substrate for Foundation-Model Software Agents},
  author={Zhong, Hailin and Zhu, Shengxin},
  journal={arXiv preprint arXiv:2605.13357},
  year={2026}
}

@article{chen2025cudallm,
  title = {{CUDA-LLM}: {LLMs} Can Write Efficient {CUDA} Kernels},
  author={Chen, Wentao and Zhu, Jiace and Fan, Qi and Ma, Yehan and Zou, An},
  journal={arXiv preprint arXiv:2506.09092},
  year={2025}
}

@article{zhang2025cudaforge,
  title = {{CudaForge}: An Agent Framework with Hardware Feedback for CUDA Kernel Optimization},
  author={Zhang, Zijian and Wang, Rong and Li, Shiyang and Luo, Yuebo and Hong, Mingyi and Ding, Caiwen},
  journal={arXiv preprint arXiv:2511.01884},
  year={2025}
}

@article{li2026harbor,
  title = {{HARBOR}: A Harness Framework for Agentic Robot Reinforcement Learning},
  author={Li, Zechu and Jin, Yufeng and Liu, Xiaoyang and Liu, Puze and Prasad, Vignesh and D'Eramo, Carlo and Chalvatzaki, Georgia},
  journal={arXiv preprint arXiv:2606.08610},
  year={2026}
}

@article{jin2026nautilus,
  title = {{Nautilus}: From One Prompt to Plug-and-Play Robot Learning},
  author = {Yufeng Jin and Jianfei Guo and Xiaogang Jia and Yu Deng and Zechu Li and Han Liu and Weiran Liao and Vignesh Prasad and Mathias Franzius and Gerhard Neumann and Georgia Chalvatzaki},
  journal = {arXiv preprint arXiv:2605.11665},
  year = {2026},
}

@article{chen2026showharness,
  title={{Show-Harness}: Just a {VLM} Agent Can Play Robots},
  author={Chen, Yanzhe and Bai, Zechen and Cao, Zhijun and Zeng, Wenzheng and Lin, Kevin Qinghong and Lin, Yiqi and Liang, Guoqiang and Ma, Kevin Yuchen and Huang, Qiming and Shou, Mike Zheng},
  journal={arXiv preprint arXiv:2609.10522},
  year={2026}
}

@article{li2026sol,
  title = {{Sol Video Inference Engine}: Agent-Native Full-Stack Acceleration Framework for Efficient Video Generation},
  author = {Yitong Li and Junsong Chen and Haopeng Li and Haozhe Liu and Jincheng Yu and Ligeng Zhu and Ping Luo and Song Han and Enze Xie},
  journal = {arXiv preprint arXiv:2606.23743},
  year = {2026}
}

@article{wang2026openwam,
  title   = {{OpenWAM}: An Open, Modular Exploration Towards Systematic World-Action Model Pretraining},
  author  = {Yuran Wang and Siqiao Huang and Mingleyang Li and Chenhao Zhang and Jiaqi Liang and Weiyang Jin and Yue Chen and Xuemin Chi and Donghao Zhou and Qize Yu and Yu-Kai Wang and Yuhan Rui and Shenzhe Yao and Zhen Yuan and Zhenhao Shen and Kefei Zhu and Zijie Zhu and Ning Gao and Xiaowei Chi and Guanqi He and Shanghang Zhang and Hao Dong and Lin Shao and Hang Zhao},
  year    = {2026},
  journal = {arXiv preprint arXiv: 2609.07398}
}

@inproceedings{yang2026robolab,
    author    = {Xuning Yang and Rishit Dagli and Alex Zook and Hugo Hadfield and Ankit Goyal and Stan Birchfield and Fabio Ramos and Jonathan Tremblay},
    title     = {{RoboLab}: A High-Fidelity Simulation Benchmark for Analysis of Task Generalist Policies},
    booktitle = {Proceedings of Robotics: Science and Systems},
    year      = {2026},
    address   = {Sydney, Australia},
    month     = {July},
    doi       = {10.15607/RSS.2026.XXII.096}
}

@article{agarwal2026cosmos3,
  title={Cosmos 3: Omnimodal world models for physical {AI}},
  author={Agarwal, Niket and Ali, Arslan and Allen, Jon and Antolini, Martin and Aubame, Adeline and Azzolini, Alisson and Bai, Junjie and Bala, Maciej and Balaji, Yogesh and Bapst, Josh and others},
  journal={arXiv preprint arXiv:2606.02800},
  year={2026}
}

@article{lee2026meta,
  title={{Meta-Harness}: End-to-end optimization of model harnesses},
  author={Lee, Yoonho and Nair, Roshen and Zhang, Qizheng and Lee, Kangwook and Khattab, Omar and Finn, Chelsea},
  journal={arXiv preprint arXiv:2603.28052},
  year={2026}
}
\end{document}